# Optimal Transport Image Representation and Deep Covariance Alignment (CORAL) for Control Valve Stiction Detection

**Seshu K. Damarla**

*Department of Chemical and Materials Engineering, University of Alberta, Edmonton, Alberta, Canada (damarla@ualberta.ca)*

**Abstract**: Control valve stiction is a common cause of unwanted oscillations and poor control-loop performance in industrial processes. Data-driven methods can automatically detect stiction, but models trained purely on simulated data often struggle to generalize to real industrial control loops due to domain shift. To bridge this gap, this work propose a novel stiction detection methodology that combines optimal transport (OT) imaging technique with deep correlation alignment (Deep CORAL) algorithm. Closed loop signals: controller output and process variable are converted into two-dimensional OT images. These images capture the dynamic behaviour of control loops. The proposed methodology includes a convolutional neural network (CNN) encoder (or feature extractor) trained to learn domain-invariant representations by optimizing a combined objective: a cross-entropy loss on labeled simulation data and a Deep CORAL (covariance-alignment) loss between simulation data and unlabeled target-domain industrial data. Downstream classifiers trained on the domain-invariant target features were evaluated on an independent test set of 20 benchmark loops from industrial stiction data benchmark. The proposed methodology successfully diagnosed 18 out of the 20 loops and achieved100% recall across all 13 stiction cases, an accuracy of 90.00% and an F1-score of 92.86%. Compared to standard baseline approach (hand-crafted features-based method), the proposed methodology significantly mitigates domain shift, providing robust, highly reliable stiction detection for real-world industrial control loops.

## 1. INTRODUCTION

Control valve stiction can introduce sustained oscillations in feedback control loops [1–3]. These oscillations can reduce product quality, increase energy consumption, accelerate valve wear, and lower plant productivity. Control-loop oscillations may also result from poor controller tuning, sensor or actuator faults, loop interactions, poor process design, and external disturbances. Although invasive tests can detect stiction, they require the valve to be taken out of operation and need expert support. Therefore, non-invasive methods using routine controller output (OP) and process variable (PV) data are preferred [4]. AI and machine-learning techniques, such as fuzzy clustering, nonlinear principal component analysis, neural networks, and CNN-based pattern recognition, have further improved automatic stiction detection and separation from other oscillation causes [5,6].

Transforming control-loop time series signals into images can reveal temporal, lagged, and cross-signal patterns that may be difficult to identify directly from long OP and PV sequences. It also allows computer vision algorithms to learn useful features automatically. Hajimehdigholi et al. converted OP and PV signals into Toeplitz matrices and used a CNN–LSTM model to learn both spatial and sequential patterns for stiction detection [7]. Memarian et al. used Gramian Angular Field images with a stacked autoencoder to identify poorly tuned controllers [8], while their earlier work converted OP and PV signals into Markov Transition Field images and used a CNN with transfer learning for stiction detection [9]. Damarla et al. used the visual geometry of the OP–ΔPV phase plot and fitted a sigmoid function to its characteristic S-shaped pattern [10]. Krishnan et al. used actual camera images and edge detection to obtain non-contact valve-position feedback; however, this approach requires additional imaging hardware and differs from methods based only on routine closed-loop signals [11]. Other widely used time-series imaging methods include Gramian Angular Summation and Difference Fields, Markov Transition Fields [12], recurrence plots [13], Fourier spectrograms [14], wavelet scalograms [15], and Toeplitz matrices. These methods represent angular correlations, state-transition probabilities, recurrent states, frequency content, or lagged samples. In comparison, the optimal transport imaging technique measures sliced-Wasserstein distance between multivariable signal windows. It therefore compares the complete distributions of OP, PV, their rate-of-change relationship, and tracking error (OP-PV) rather than relying only on pointwise amplitudes, predefined frequency bands, quantized states, or a specific waveform. This representation can capture changes in distribution shape, spread, location, and is less dependent on exact sample-to-sample alignment, and does not require the amplitude discretization used by MTF or the recurrence threshold used in recurrence plots. The sliced-Wasserstein approximation also reduces the computational cost of full optimal transport by using one-dimensional projections [16,17]. Thus, OT imaging offers a more suitable representation for separating stiction-related stick–slip behaviour from other oscillatory conditions, although its performance still depends on the selected window size and number of projections.

Based on these observations, this work proposes a novel non-invasive valve-stiction detection framework combining the optimal transport imaging technique with Deep covariance alignment (CORAL) domain adaptation algorithm. The Deep CORAL reduces the difference between simulated data and industrial data by aligning their feature covariance structures [18]. A baseline method based on hand-crafted features is developed to compare its performance with that of the proposed method.

## 2. PRELIMINARIES

### *2.1. Control Loop Behavior under Valve Stiction*

In industrial processes, flow of fluids is regulated by final control elements such as control valves. In the ideal case of zero static friction, a control valve accurately executes the controller output (OP) generated by a feedback controller. As a result, the valve position smoothly changes with OP, ensuing in a nearly linear relationship between them. Conversely, the manifestation of valve stiction turns the linear relationship into a nonlinear relationship as shown in Figure 1 [1-3]. Due to static friction between valve stem and packing, a sticky control valve does not quickly act on small changes in OP, this behaviour is termed stiction band. When the accumulated change in OP is large enough, the control valve overcomes the static friction and abruptly moves to a new position. The abrupt jump in valve position is called slip-jump. After the slip-jump, the valve stem continues (this is called moving phase) to move until OP changes its direction. While the control valve is in operation, these three phases occur repeatedly, thus, the nonlinear relationship is developed. During these phases, process variable (PV) deviates from its desired value and the controller tries to reduce error by adjusting OP. As the three phases happen repetitively, both PV and OP oscillate hence they carry important information about the presence of stiction (Figure 2). It is easy to detect stiction if actual valve position (i.e. valve output) is available, but it is rarely measured in practice. For this reason, non-intrusive stiction detection methodologies depend on PV and OP.

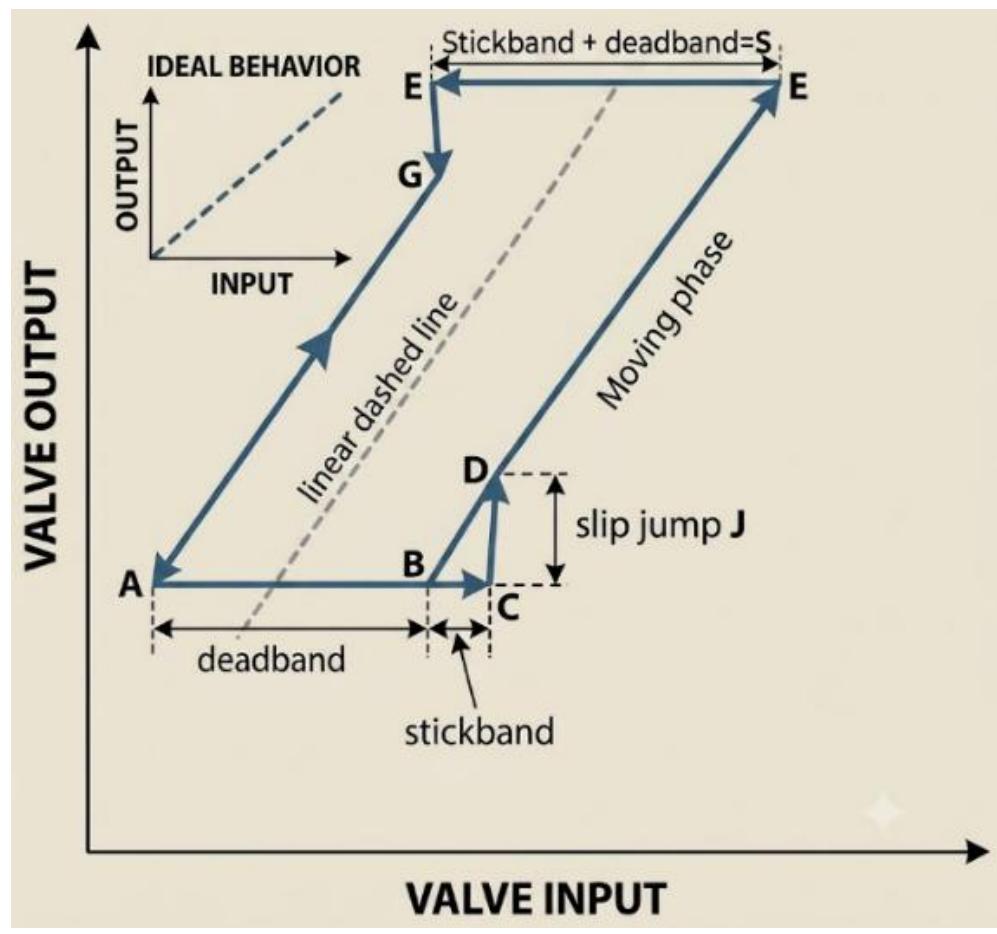

Figure 1. Behavior of sticky control valve.

### *2.2. Optimal Transport Image Generation*

Optimal transport computes resemblance between two probability distributions through the determination of minimum transportation cost required to transform one distribution into another [16, 17]. OT is utilized to build an image from time series signals, capturing relationships between different segments of the same signal in preference to representing the entire signal directly. The procedure to construct OT image from PV and OP of a control loop is explained in the following.

- Obtain PV and OP from a feedback control loop.
- Normalize PV and OP using min-max normalization technique.
- Calculate the first order derivatives of the normalized PV and OP signals.
- Compute ${dOP}/{(dPV + \epsilon)}$, where $\epsilon$ is a small positive constant to avoid division by zero.
- Compute control error: $(OP - PV)$
- Construct a feature vector
  $$x(t) = \left[PV \; OP \; {dOP}/{(dPV + \epsilon)} \; (OP - PV)\right]$$
- Divide $x(t)$ into overlapping windows using a predefined window length and stride.
- Generate multiple random unit vectors for sliced optimal transport projects.
- Project the feature vectors of each window in every random direction.
- Compute one-dimensional Wasserstein distance between every pair of projected windows.
- Compute mean of Wasserstein distances over all projection directions to obtain sliced Wasserstein distance. Repeat this step for all window pairs to construct an OT distance matrix.
- Normalize the OT matrix to $[0,1]$.
- Apply Gaussian smoothing to suppress noise and produce a smoother distance map while preserving the overall structural patterns.
- Interpret the normalized and smoothed distance matrix as an image, where each matrix element corresponds to the intensity of one image pixel.
- The OT image represents the dynamics of the given control loop.

To the best of my knowledge, this work introduces the first application of optimal transport imaging within the domain of control loop performance monitoring.

## 3. THE PROPOSED METHODOLOGY

In this section, the proposed framework is delineated comprehensively. Figure 2 shows the overall architecture of the proposed methodology.

The source domain dataset (i.e. simulation dataset) is denoted by

$$\mathcal{D}_s = \{(I_s^i, y_s^i)\}_{i=1}^{n_s}, \quad (1)$$

where $I_s^i$ signifies the OT image created from the $i$-th simulated control loop, $y_s^i \in \{0,1\}$ is the label for this image, $n_s$ is the number of source-domain images.

If the $i$-th control loop has a sticky control valve, then $y_s^i = 1$ otherwise $y_s^i = 0$.

The target-domain training set is expressed as

$$\mathcal{D}_t = \{I_t^j\}_{j=1}^{n_t}, \quad (2)$$

where $I_t^j$ is the OT image attained from the $j$-th ISDB control loop.

The target-domain class labels are not required to train Deep CORAL. The target images are utilized solely to minimize the discrepancy between the industrial and the source-domain feature distributions.

The same CNN encoder is used to process the source and the target images, so the two domains share the same encoder parameters. The encoder comprises three convolution blocks, a global average pooling layer and a fully connected layer. Each convolution block contains a two-dimensional convolutional layer, a batch normalization layer, a ReLU activation layer and a maximum pooling layer. A source-domain image ($I_s^i$) is fed to the CNN encoder. The image is processed first in the first convolutional block, in which the convolutional layer extracts spatial features from the image. The features are normalized in the batch normalization layer to have zero mean and unit variance. The normalized features are passed through the ReLU activation layer. The output of the ReLU activation layer undergoes dimensionality reduction in the maximum pooling layer. The output of the first convolutional block is again processed in the subsequent convolution blocks. The output of the last convolutional block contains multiple feature maps and is further processed in the global average pooling layer, which computes the average value of each feature map to generate a compact feature representation for the subsequent fully connected layer. The fully connected layer produces feature representation of the source-domain image.

$$h_s^i = f_\theta(I_s^i), \quad (3)$$

where $f_\theta(\cdot)$ is the CNN encoder and $\theta$ represents the trainable parameters.

In the same way, feature representation of a target-domain image ($I_t^j$) can be obtained:

$$h_t^j = f_\theta(I_t^j). \quad (4)$$

Feature embeddings for the rest of the source-domain and the target-domain images can be obtained in the same fashion.

To impart supervised information to the encoder (which allows the encoder to learn features that can distinguish stiction from non-stiction), a temporary classification head is added to the output of the encoder. This classification layer maps the source-domain feature vectors to two logits belonging to stiction and non-stiction.

$$O_s^i = g_\emptyset(h_s^i), \quad (5)$$

where $g_\emptyset(\cdot)$ represents the classification layer and $\emptyset$ contains its learnable parameters.

A softmax function is used to convert the logits into class probabilities.

$$p_{ic} = \frac{exp(O_{ic})}{\sum_{k=1}^{2} exp(O_{ik})}, c \in \{0,1\}. \quad (6)$$

The parameters of the encoder and the classification head are learned by minimizing the class-weighted cross-entropy loss function given below.

$$\mathcal{L}_{CE} = -\frac{1}{n_s}\sum_{i=1}^{n_s}\sum_{c=1}^{2} w_c \mathbb{I}(y_s^i = c) log(p_{ic}), \quad (7)$$

where $w_c$ is the weight assigned to class $c$, and $\mathbb{I}(\cdot)$ is the indicator function. The class distribution of the source-domain images is utilized to compute the class weights. The class-weighted cross entropy function deals with the class imbalance in the source-domain.

Since the simulation data cannot fully replicate actual plant operations, the distribution of the source-domain features may be dissimilar to the distribution of the target-domain features. To reduce this source-to-target domain difference, this work introduces deep covariance alignment (Deep CORAL).

Let us assume that $H_s$ is a mini batch drawn from the source-domain images and $H_t$ from the target-domain images.

$$H_s = \begin{bmatrix} (h_s^1)^T \\ \vdots \\ (h_s^{m_t})^T \end{bmatrix} \epsilon \mathbb{R}^{m_s \times d}, \quad (8)$$

$$H_t = \begin{bmatrix} (h_t^1)^T \\ \vdots \\ (h_t^{m_t})^T \end{bmatrix} \epsilon \mathbb{R}^{m_t \times d}, \quad (9)$$

where $m_s$ and $m_t$ are the sizes of the source and target mini batches, respectively, and $d$ is the dimension of the feature vector.

The centered covariance matrices are determined as

$$C_s = \frac{1}{m_s - 1}(H_s - \widetilde{H}_s)^T(H_s - \widetilde{H}_s), \quad (10)$$

$$C_t = \frac{1}{m_t - 1}(H_t - \widetilde{H}_t)^T(H_t - \widetilde{H}_t), \quad (11)$$

where $\widetilde{H}_s$ and $\widetilde{H}_t$ are the feature means of $H_s$ and $H_t$, respectively.

The Deep CORAL loss is defined as

$$\mathcal{L}_{CORAL} = \frac{1}{4d^2}\|C_s - C_t\|_F^2, \quad (12)$$

where $\|\cdot\|_F$ is the Frobenius norm.

The covariance of the source-domain features is brought close to that of the target-domain features by minimizing the Deep CORAL loss function. Accordingly, the CNN encoder is forced to learn domain-invariant (or domain-agnostic) representations. Eq. (7) and Eq. (12) are combined to form the final loss function shown in Eq. (13) that is minimized to train the encoder.

$$\mathcal{L}_{total} = \mathcal{L}_{CE} + \lambda\mathcal{L}_{CORAL}, \quad (13)$$

where $\lambda$ controls the contribution of domain alignment.

The following strategy for $\lambda$ is used during training.

$$\lambda(e) = \lambda_{max}\left[\frac{2}{1 + exp(-10p_e)} - 1\right], p_e = \frac{e}{E-1}, \quad (14)$$

where $e$ is the current epoch, $E$ is the total number of epochs, and $\lambda_{max} = 1$.

For each training iteration, one batch of labeled simulation images and one batch of unlabeled ISDB training images are processed in the shared encoder. Simulation images and their respective labels are utilized for computing the weighted cross-entropy while the ISDB feature vectors and the simulation feature vectors are used to calculate the CORAL loss. The total loss is decreased to jointly update the parameters of the encoder and the temporary classification head. After the completion of training, the shared encoder is retained, and the temporary classification head is removed. The weights of the trained shared encoder are not further updated. The ISDB training images are passed through the CNN encoder to extract domain-adapted features. These features are standardized using z-score normalization. The normalized features are related to the class labels of the ISDB training images by the help of logistic regression. The probability that a given control loop has a sticky valve is determined as

$$P(y_i = 1|z_i) = \frac{1}{1+exp[-(w^T z_i+b)]}, \quad (15)$$

where $w$ is the logistic regression coefficient vector and $b$ is the intercept.

The optimum values of $w$ and $b$ are obtained by minimizing the following regularized binary cross-entropy objective function.

$$\mathcal{J} = -\sum_{i=1}^{N}[y_i log(p_i) + (1-y_i) log(1-p_i)] + \frac{1}{2C}\|w\|_2^2, \quad (16)$$

where $C$ balances regularization strength.

The hyperparameters $C$ and the use of balanced class weights are selected using cross-validation on the ISDB training features. To avoid data leakage, the decision threshold ($\tau$) is likewise computed using only out-of-fold probabilities from the ISDB training set. For the optimized threshold, the proposed methodology predicts a class label for a given image as per Eq. (17).

$$\hat{y}_i = \begin{cases} 1, & P(y_i = 1|z_i) \geq \tau, \\ 0, & P(y_i = 1|z_i) < \tau. \end{cases} \quad (17)$$

It is to be noted that the first and second stages are trained one by one. First, the CNN encoder and the temporary classification head are trained together using the source-domain labeled images and the ISDB training images without labels. The trained classification head is disconnected, and the trained CNN encoder is kept. The ISDB training images are fed to the CNN encoder to get features that are, in turn, mapped to the class labels of the training images. The ISDB test images are not used either in minimizing the total loss (cross-entropy plus Deep CORAL loss) or determining the unknowns of the final logistic regression classifier.

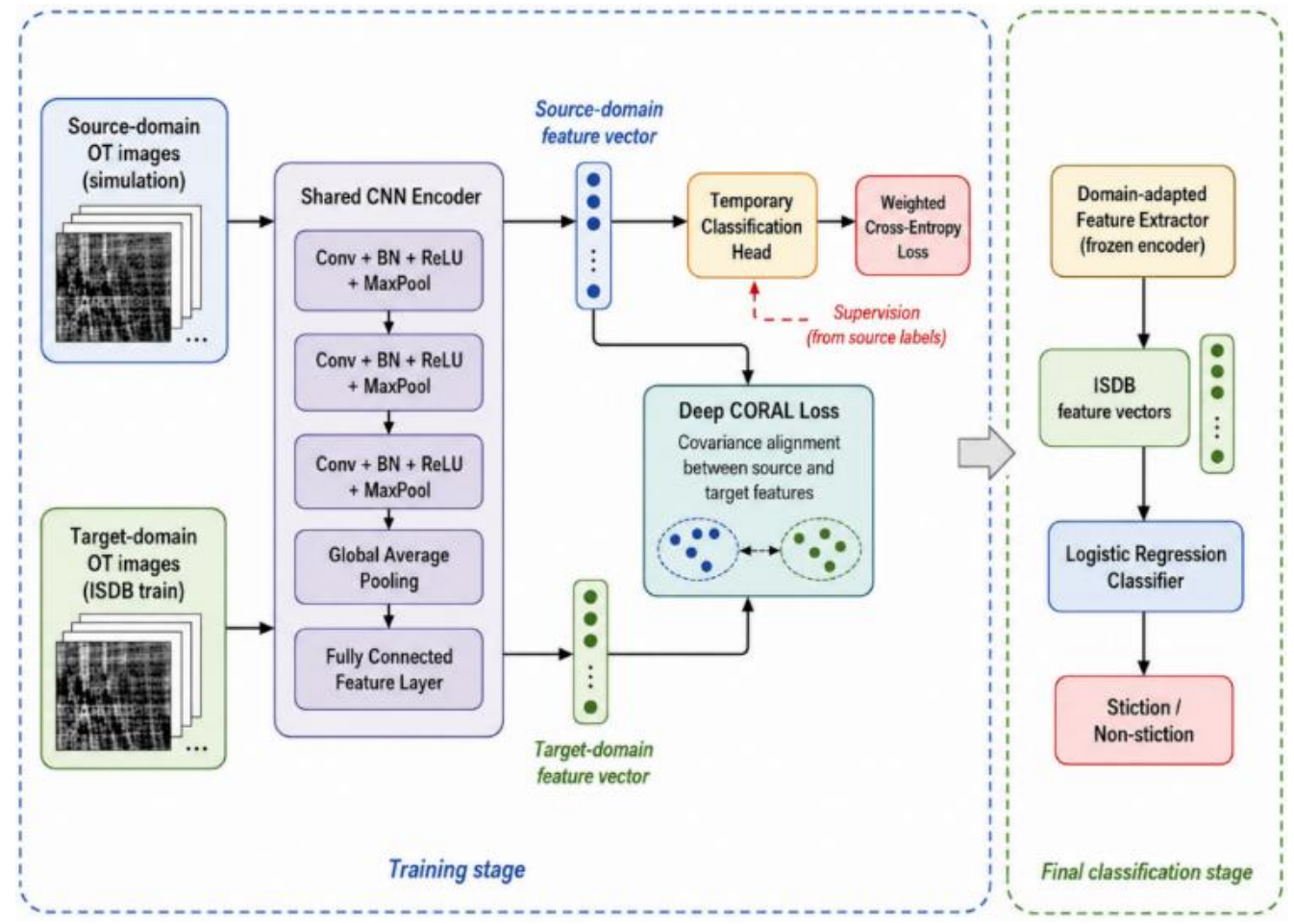

Figure 2. Configuration of proposed methdology

# 4. EXPERIMENTAL STUDY

## *4.1. Simulation and Industrial Data*

In the experimental study, the simulation database was used as the source-domain and the ISDB database as the target-domain. The simulation database was created by simulating the feedback control loop shown in Figure 3.

For the simulation database to have a wide variety of oscillations under diverse scenarios, the concentration control loop (self-regulating process) shown in Eq. (18) and the level control loop (integrating process) given in Eq. (19).

$$G_p(s) = \frac{3e^{-10s}}{10s+1}, G_c(s) = K_c\left(1 + \frac{1}{\tau_I s}\right) \quad (18)$$

$$G_p(s) = \frac{1}{s}, G_c(s) = K_c\left(1 + \frac{1}{\tau_I s}\right) \quad (19)$$

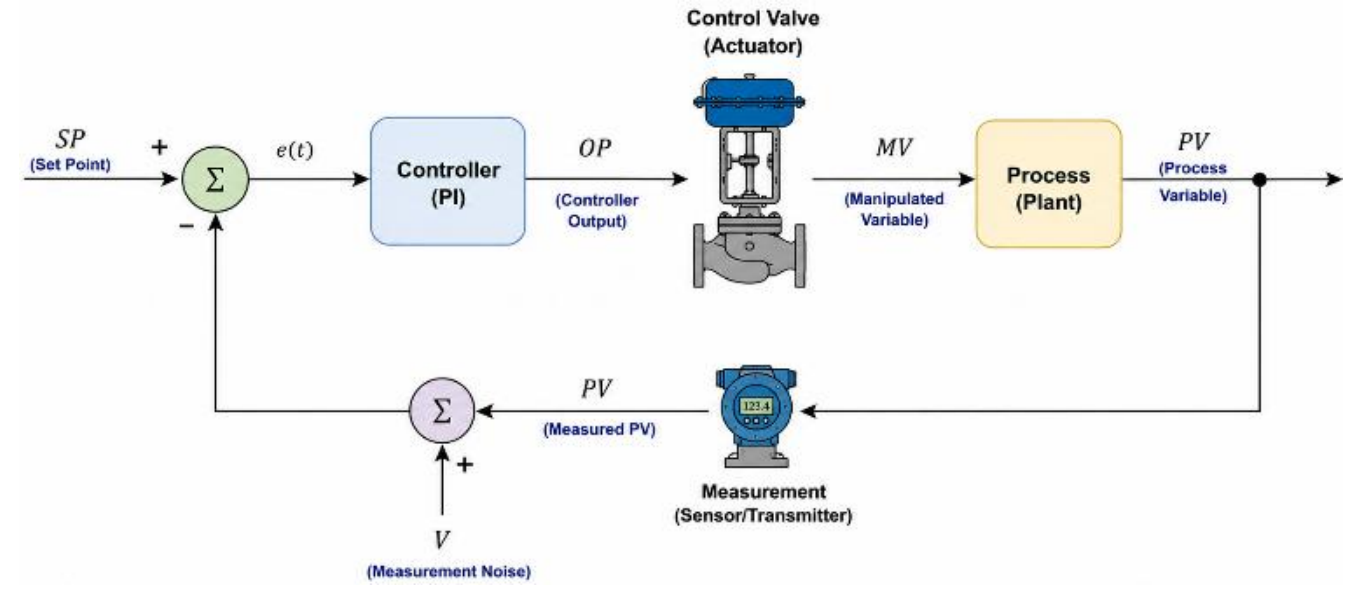

Figure 3. Simulated feedback control loop.

To simulate the behavior of a sticky control valve in the closed loop, data-based stiction model proposed by Shoukat Choudhury et al. (2005) was used in the present work. SIMULINK was employed to run simulations under different stiction and non-stiction conditions (tightly tuned controller & oscillatory disturbances) provided in Table 1. In Table I, $S$ and $J$ indicate the stiction band and the slip-jump parameters of the stiction model. $A$ and $F$ denote the amplitude and the frequency of a sinusoidal signal used to create oscillatory disturbances. The same ranges for $K_c$ and $\tau_I$ were used for both control loops. In each simulation, PV was corrupted with Gaussian noise with zero mean and variance V. 2721 simulations were run to create 1495 stiction loops and 1226 non-stiction loops. The industrial database contained 79 control loops gathered from six industrial sectors: buildings, chemicals, pulp and paper, power generation, mining and metals. The ISDB control loop data was converted into OT images that were randomly split into training and test sets. Table 2 provides the information for the source-domain and the target-domain images. Table 3 bestows the configuration and training parameters of the OT image generation and the proposed method.

Table 1. Parameter values/ranges for stiction and non-stiction conditions

| Scenario | Parameter | Value(s)/range |
|---|---|---|
| Stiction | $S$ | $[0.5: 0.25: 10]$ |
| | $J$ | $[0.1: 0.25: 5]$ |
| | $V$ | 0.01 |
| Tightly tuned controller | $K_c$ | $[0.1: 0.01: 0.3]$ |
| | $\tau_I$ | $[0.01: 0.01: 0.27]$ |
| | $V$ | 0.01 |
| Oscillatory disturbances | $A$ | $[1,1.5,2,2.5]$ |
| | $F$ | $[0.01: 0.01: 0.27]$ |
| | $V$ | 0.01 |

Table 2. Source-domain and target-domain datasets

| Dataset | Domain | Purpose | NSL | NNSL |
|---|---|---|---|---|
| Simulation | Source | DCT | 1495 | 1226 |
| Training ISDB | Target | AL & CT | 22 | 37 |
| Test ISDB | Target | FT | 13 | 7 |

DCT – Deep CORAL training, AL & CT – alignment and final classifier training, FT – final testing, NSL – no. of stiction loops, NNSL – no. of non-stiction loops.

Table 3. OT image generation and CNN encoder training details

| Parameter | Value |
|---|---|
| Image size | 50×50 |
| Window length | 20 |
| Stride | 10 |
| Sliced projections | 10 |
| Maximum windows | 50 |
| CNN feature dimension | 64 |
| Batch size | 32 |
| Training epochs | 100 |
| Optimizer | Adam |
| Learning rate | |
| Weight decay | |
| Maximum coral weight | |

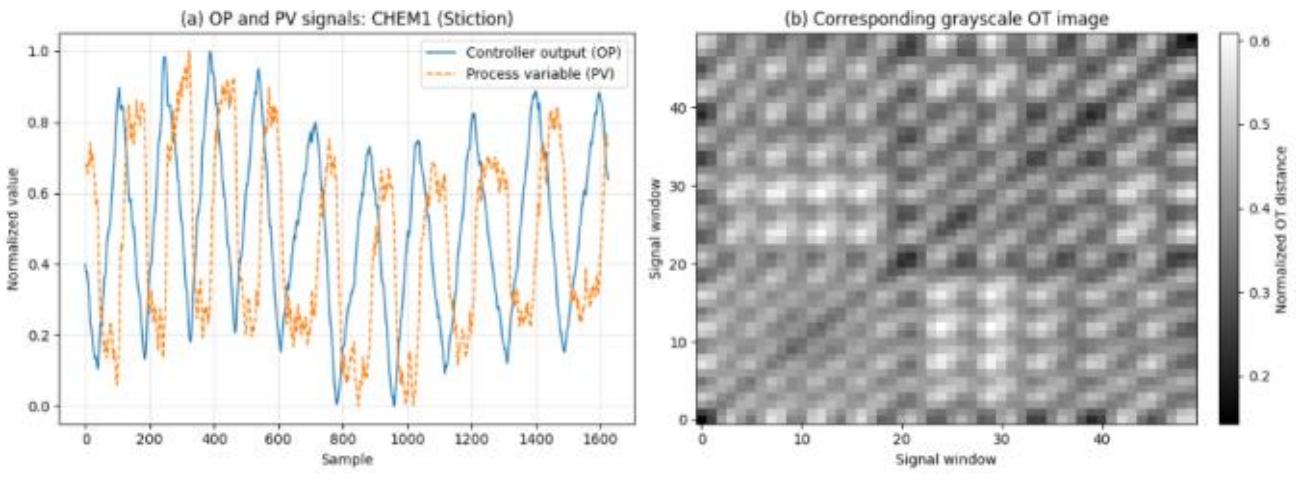

Figure 4. OT image representation for stiction-induced oscillations.

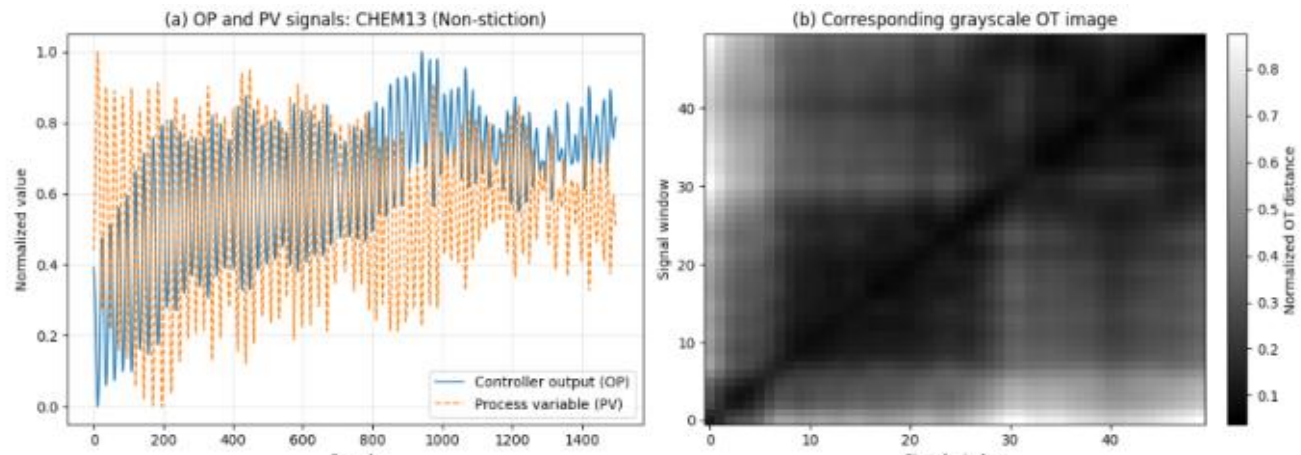

Figure 5. OT image representation for oscillations caused by a non-stiction condition.

Figures 4 and 5 provide representative OT images for stiction and non-stiction examples, respectively.

### 4.2. *Baseline Method*

A baseline method was developed using hand-crafted features extracted from the ISDB training images. The hand-crafted features obtained by adopting the traditional image feature extraction methods given in Table 4. The HOG, LBP, GLCM and the statistical features were concatenated to form a single vector for each ISDB image. The feature vectors were normalized and then used, along with the class labels, to develop a binary classifier using $L_2$-regularized logistic regression algorithm. Optuna was used to select the regularization parameter and the use of class-balanced weights. The classifier parameters and hyperparameters were obtained using training-only cross-validation. For an acceptable comparison, the baseline method used the same OT image generation parameters, the same ISDB training images and was evaluated using the same ISDB test images. The classification performance of the baseline method was quantified using the same metrics as the proposed method.

Table 4. Traditional image feature extraction methods

| Method | Features |
|---|---|
| HOG | Edges, shapes and gradient directions |
| LBP | Local texture patterns |
| GLCM | Texture properties: contrast, homogeneity, energy, correlation |
| Statistical features | Standard deviation, mean, minimum, maximum, median, skewness, kurtosis, entropy |

HOG - histogram of oriented gradients, LBP - local binary pattern, GLCM - gray-level co-occurrence matrix

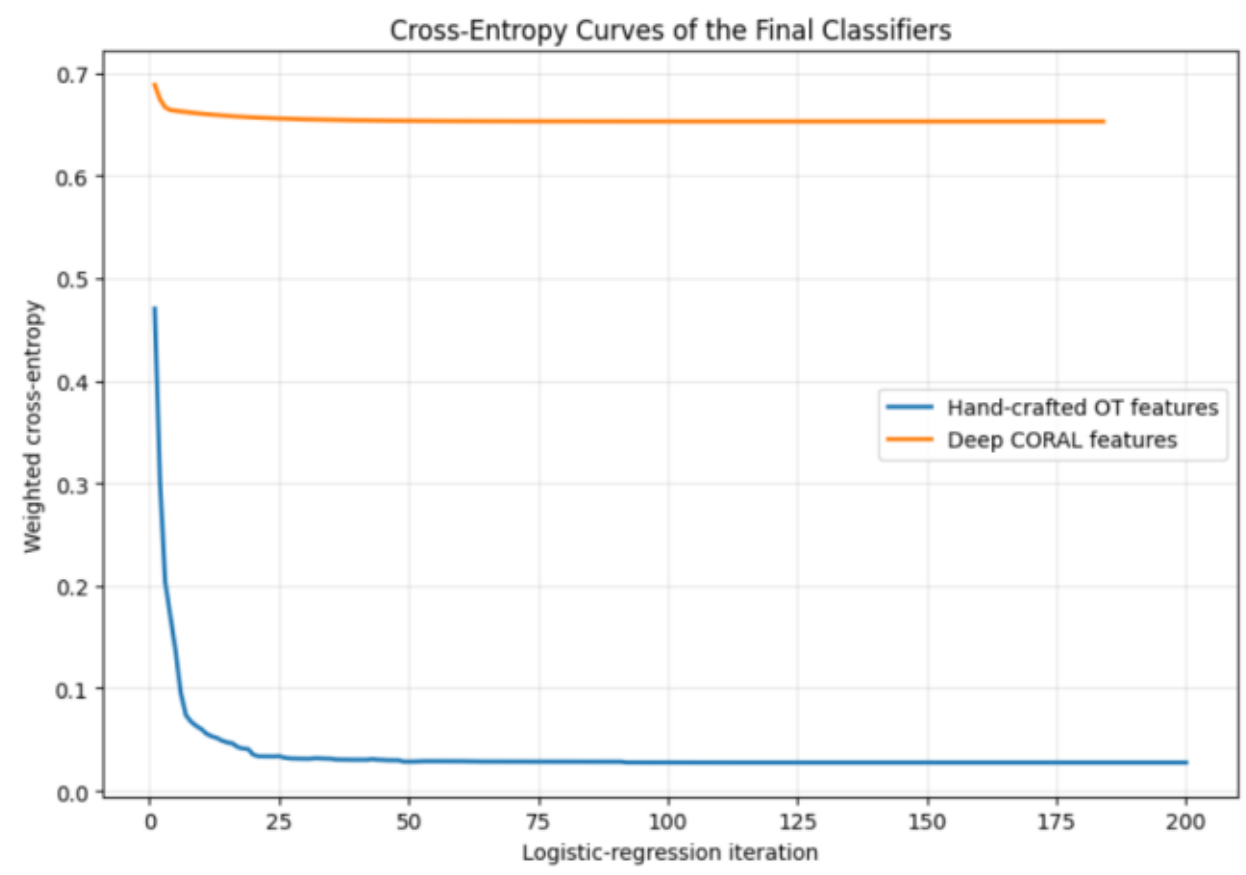

Figure 6. Learning curve of baseline classifier and the final classifier of the proposed method.

## 5. RESULTS AND DISCUSSIONS

Figure 6 displays the performance of the final classifier of the proposed method and that of the baseline method. The proposed method was evaluated on the 20 unseen industrial control loops from the ISDB, including 13 stiction and seven non-stiction loops. These loops were not used for domain adaptation, classifier training, hyperparameter tuning, or threshold selection. The final diagnosis was obtained by testing the final classifier on the 64-dimensional domain-adapted features extracted from the test OT images. As shown in Table 5, the baseline method correctly diagnosed 13 of the 20 loops while the proposed method correctly diagnosed 18 loops. The baseline method missed the four stiction loops (CHEM2, CHEM24, CHEM29, and MIN1) and incorrectly classified the three non-stiction loops as stiction. In contrast, the proposed method correctly diagnosed all 13 stiction loops and produced no false negatives. Its only errors were PAP4 and PAP9, which were non-stiction loops classified as stiction. The proposed method performed especially well on the chemical and mining loops. It correctly diagnosed all 14 chemical loops and the mining loop MIN1. However, two errors occurred in the pulp-and-paper loops. These false positives may be caused by non-stiction oscillations that have OP–PV patterns similar to those produced by valve stiction.

Table 6 shows that the proposed method achieved an accuracy of 0.9000, precision of 0.8667, recall of 1.0000, and F1-score of 0.9286. The baseline achieved an accuracy of 0.6500, precision of 0.7500, recall of 0.6923, and F1-score of 0.7200. Thus, the proposed method improved accuracy by 25 percentage points and increased the F1-score by about 20.86 percentage points.

Table 5. Results for test ISDB loops

| Loop | Actual malfunction | Baseline diagnosis | PM diagnosis |
|---|---|---|---|
| CHEM1 | Stiction | Stiction | Stiction |
| CHEM2 | Stiction | NS | Stiction |
| CHEM3 | NS | Stiction | NS |
| CHEM6 | Stiction | Stiction | Stiction |
| CHEM10 | Stiction | Stiction | Stiction |
| CHEM11 | Stiction | Stiction | Stiction |
| CHEM12 | Stiction | Stiction | Stiction |
| CHEM13 | NS | NS | NS |
| CHEM14 | NS | NS | NS |
| CHEM16 | NS | Stiction | NS |
| CHEM23 | Stiction | Stiction | Stiction |
| CHEM24 | Stiction | NS | Stiction |
| CHEM29 | Stiction | NS | Stiction |
| CHEM32 | Stiction | Stiction | Stiction |
| PAP2 | Stiction | Stiction | Stiction |
| PAP4 | NS | NS | Stiction |
| PAP5 | Stiction | Stiction | Stiction |
| PAP7 | NS | Stiction | NS |
| PAP9 | NS | NS | Stiction |
| MIN1 | Stiction | NS | Stiction |

PM – proposed method, NS – non-stiction

Table 6. Performance quantification.

| Method | Accuracy | Precision | Recall | F1-score |
|---|---|---|---|---|
| Baseline | 0.6500 | 0.7500 | 0.6923 | 0.7200 |
| PM | 0.9000 | 0.8667 | 1.0000 | 0.9286 |

The baseline produced nine true positives, four true negatives, three false positives, and four false negatives. The proposed method produced 13 true positives, five true negatives, two false positives, and no false negatives. This confirms that the proposed method improved both stiction detection and non-stiction recognition. The improvement is due to the combined use of OT-image representation, domain adaptation, feature extraction, and classification. Overall, the results show that the proposed method provides more reliable stiction detection than the baseline method.

## 6. CONCLUSIONS

This study developed a domain-adaptation-based methodology for detecting control valve stiction using the optimal transport image technique. The proposed framework converted OP and PV into images and extracts domain-adapted features for classifying industrial loops as stiction or non-stiction. The method was trained using the simulated data together with a limited amount of industrial data and was evaluated on the independent ISDB test loops. The proposed method correctly diagnosed 18 of the 20 test loops and achieved an accuracy of 0.9000, precision of 0.8667, recall of 1.0000, and F1-score of 0.9286. In comparison, the baseline method achieved an accuracy of 0.6500 and an F1-score of 0.7200. The proposed method detected all 13 stiction loops and produced no false-negative diagnoses. It also correctly identified several stiction loops that were missed by the baseline method. These results demonstrate that the proposed framework can reduce the difference between simulated and industrial data and improve the reliability of valve-stiction diagnosis.

The only misclassifications observed were two false positives in non-stiction pulp-and-paper loops, demonstrating that some non-stiction oscillations may have patterns like those caused by valve stiction. To build upon these findings, the planned future work will focus on enhancing feature discrimination between valve stiction and non-stiction oscillations, as well as validating the framework across broader industrial processes and diverse operating regimes. Overall, these results confirm that combining OT imaging with domain adaptation provides a highly dependable, zero-false-negative diagnostic tool for control performance monitoring.

## REFERENCES

[1] M. A. Daneshwar and N. Mohd Noh, "Detection of stiction in flow control loops based on fuzzy clustering," *Control Engineering Practice*, vol. 39, pp. 23–34, 2015.

[2] A. S. R. Brásio, A. Romanenko, and N. C. P. Fernandes, "Detection of stiction in level control loops," *IFAC-PapersOnLine*, vol. 48, no. 8, pp. 421–426, 2015.

[3] J. W. V. Dambros, M. Farenzena, and J. O. Trierweiler, "Data-based method to diagnose valve stiction with variable reference signal," *Industrial & Engineering Chemistry Research*, vol. 55, pp. 10316–10327, 2016.

[4] S. K. Damarla, X. Sun, F. Xu, A. Shah, J. Amalraj, and B. Huang, "Practical linear regression-based method for detection and quantification of stiction in control valves," *Industrial & Engineering Chemistry Research*, vol. 61, pp. 502–514, 2022.

[5] W. K. Teh, H. Zabiri, Y. Samyudia, S. S. Jeremiah, B. Kamaruddin, A. A. A. Mohd Amiruddin, and N. M. Ramli, "An improved diagnostic tool for control valve stiction based on nonlinear principal component analysis," *Industrial & Engineering Chemistry Research*, vol. 57, pp. 11350–11365, 2018.

[6] D. Zheng, X. Sun, S. K. Damarla, A. Shah, J. Amalraj, and B. Huang, "Valve stiction detection and quantification using a K-means clustering-based moving-window approach," *Industrial & Engineering Chemistry Research*, vol. 60, pp. 2563–2577, 2021.

[7] Z. Hajimehdigholi, S. K. Damarla, and B. Huang, "A Novel Toeplitz Matrix and CNN-LSTM Based Method for Identifying Control Valve Stiction," *Industrial & Engineering Chemistry Research*, vol. 64, no. 34, pp. 16757–16769, 2025.

[8] A. Memarian, S. K. Damarla, A. Memarian, and B. Huang, "Detection of Poor Controller Tuning with Gramian Angular Field (GAF) and StackAutoencoder (SAE)," *Computers & Chemical Engineering*, vol. 185, Article 108652, 2024.

[9] A. Memarian, S. K. Damarla, and B. Huang, "Control Valve Stiction Detection Using Markov Transition Field and Deep Convolutional Neural Network," *The Canadian Journal of Chemical Engineering*, vol. 101, no. 11, pp. 6114–6125, 2023.

[10] S. K. Damarla, X. Sun, F. Xu, A. Shah, and B. Huang, "A Sigmoid Function Based Method for Detection of Stiction in Control Valves," in *2022 IEEE International Symposium on Advanced Control of Industrial Processes*, pp. 210–215, 2022.

[11] S. Krishnan, C. Thomas, D. G. Dharmaraj, Devakrishna K. V., C. S. Suryaraj, and K. N. Kailas, "Computer Vision-Based Advanced Valve Positioner," in *2022 Third International Conference on Intelligent Computing, Instrumentation and Control Technologies*, pp. 791–795, 2022.

[12] Z. Wang and T. Oates, "Imaging Time-Series to Improve Classification and Imputation," in *Proceedings of the Twenty-Fourth International Joint Conference on Artificial Intelligence*, pp. 3939–3945, 2015.

[13] J.-P. Eckmann, S. O. Kamphorst, and D. Ruelle, "Recurrence Plots of Dynamical Systems," *Europhysics Letters*, vol. 4, no. 9, pp. 973–977, 1987.

[14] J. B. Allen and L. R. Rabiner, "A Unified Approach to Short-Time Fourier Analysis and Synthesis," *Proceedings of the IEEE*, vol. 65, no. 11, pp. 1558–1564, 1977.

[15] C. Torrence and G. P. Compo, "A Practical Guide to Wavelet Analysis," *Bulletin of the American Meteorological Society*, vol. 79, no. 1, pp. 61–78, 1998.

[16] G. Peyré and M. Cuturi, "Computational Optimal Transport: With Applications to Data Science," *Foundations and Trends in Machine Learning*, vol. 11, nos. 5–6, pp. 355–607, 2019.

[17] N. Bonneel, J. Rabin, G. Peyré, and H. Pfister, "Sliced and Radon Wasserstein Barycenters of Measures," *Journal of Mathematical Imaging and Vision*, vol. 51, no. 1, pp. 22–45, 2015.

[18] B. Sun and K. Saenko, "Deep CORAL: Correlation Alignment for Deep Domain Adaptation," in *Computer Vision—ECCV 2016 Workshops*, pp. 443–450, 2016.